\documentclass[conference]{IEEEtran}
\IEEEoverridecommandlockouts
\usepackage{cite}
\usepackage{amsmath,amssymb,amsfonts}
\usepackage{graphicx}
\usepackage{textcomp}
\usepackage{xcolor}
\usepackage{hyperref}
\usepackage{algpseudocode}
\def\BibTeX{{\rm B\kern-.05em{\sc i\kern-.025em b}\kern-.08em
    T\kern-.1667em\lower.7ex\hbox{E}\kern-.125emX}}

\usepackage{amsmath, amssymb}
\usepackage{algorithm}
\usepackage{algpseudocode}

\begin{document}

\title{False Data Injection Attack Detection in Edge-based Smart Metering Networks with Federated Learning \\
}

\author{
        \IEEEauthorblockN{
        Md Raihan Uddin\IEEEauthorrefmark{1},
        Ratun Rahman\IEEEauthorrefmark{1},
		Dinh C. Nguyen\IEEEauthorrefmark{1} %Van-Dinh Nguyen\IEEEauthorrefmark{3}, 
	}
	%\IEEEauthorblockA{\IEEEauthorrefmark{1,2}School of Engineering, Deakin University, Australia \\
	\IEEEauthorrefmark{1}Department of Electrical and Computer Engineering, The University of Alabama in Huntsville, USA  \\
	 %\IEEEauthorrefmark{3}College of Engineering and Computer Science, VinUniversity, Vinhomes Ocean Park, Hanoi 100000, Vietnam \\
	 %\IEEEauthorrefmark{4}School of Engineering, Royal Melbourne Institute of Technology University, Melbourne, VIC 3000, Australia
    Emails: mu0016@uah.edu, rr0110@uah.edu, dinh.nguyen@uah.edu
    }
	\markboth{}%
	{}

\maketitle

\begin{abstract}
Smart metering networks are increasingly susceptible to cyber threats, where false data injection (FDI)  appears as a critical attack. Data-driven-based machine learning (ML) methods have shown immense benefits in detecting FDI attacks via data learning and prediction abilities. Literature works have mostly focused on centralized learning and deploying FDI attack detection models at the control center, which requires data collection from local utilities like meters and transformers. However, this data sharing may raise privacy concerns due to the potential disclosure of household information like energy usage patterns. This paper proposes a new privacy-preserved FDI attack detection by developing an efficient federated learning (FL) framework in the smart meter network with edge computing. Distributed edge servers located at the network edge run an ML-based FDI attack detection model and share the trained model with the grid operator, aiming to build a strong FDI attack detection model without data sharing. Simulation results demonstrate the efficiency of our proposed FL method over the conventional method without collaboration. 
\end{abstract}

\begin{IEEEkeywords}
Attack detection, edge computing, metering networks, federated learning
\end{IEEEkeywords}
\section{Introduction}
The smart grid has an advanced power system based on enabling reliability, efficiency, and accuracy in the power supply system. Advanced metering infrastructure (AMI) is the most important part of achieving this goal, where smart meters (SMs) are used to measure energy consumption to facilitate data analytics \cite{esmalifalak2014detecting, lin2024privacy}. Due to the open and distributed nature of AMI, SMs are vulnerable to cyber threats including false data injection (FDI) attacks that aim to modify state estimation and energy consumption recordings at SMs \cite{zhao2021federated}. This results in financial loss and poses energy recordings under threats. For example, personal information can be compromised into the meter measurement by cybercriminals for data theft and financial profit. Hence, developing an efficient solution to detect FDI attacks is of paramount importance for safe smart metering networks. 

\subsection{{Related Works}} Recently, machine learning (ML) has been employed as an efficient technique to learn and detect FDI attacks in smart grids due to its data learning and prediction abilities. For example, neural networks were employed in \cite{esmalifalak2014detecting} for estimating anomalies in meter measurements and detecting falsified system states.  Another work in \cite{he2017real} employed deep neural networks (DNNs)  to extract features of FDI attackers in AC power transmission system simulations. Further, the work in \cite{ayad2018detection} studied FDI attack with DL for transmission-line protective relays and substation automation systems.  These works typically use centralized learning, deploying FDI attack detection models at the control center, which requires data collection from local utilities like meters and transformers. \textit{However, end customers may be unwilling to share their meter measurements for attack analysis due to privacy concerns.} 

Moreover, FL has been investigated for collaborative FDI attack detection in smart grids. The authors in \cite{zhao2021federated} worked in the smart grid to detect the attack detection in the solar farms. Another author in \cite{lin2022incentive} mentioned the power grid state estimation on unknown system parameters and small decentralized data sets with strategic data owners. The author in \cite{lin2024privacy} illustrates the view on detecting the attacks on the power system focused on privacy-preserving using federated Learning. The FDI attack detection to tackle data privacy enhancing cross-silo federated learning was proposed in \cite{tran2023efficient}. \textit{Despite these research efforts, the application of FL in FDI attack detection in edge-based smart metering networks has not been investigated.}

\subsection{Motivation and Key Contributions}
Motivated by the aforementioned limitations, we present a new FDI attack detection method using an FL framework for edge-based smart metering networks. Edge computing that offers low-latency services is useful for timely running FDI attack detection models for SMs. The use of SMs has become more reliable nowadays, but there are security threats to the use of SMs. Conversely, Some basic steps have been taken to secure the network between the clients, meter, and authority. So our concern is to secure the bond between the clients, SMs, and the authority. We found some slide works, like as this research shows the existing methods for identifying FDI attacks only identify the presence of the condition;\cite{wang2020locational} they provide crucial information regarding the precise injection locations. Motivated by the latest developments in deep learning, they put forth a deep-learning-based locational detection architecture (DLLD) to accurately identify the precise locations of FDIA instantaneously. A typical bad data detector (BDD) and a convolutional neural network (CNN) are combined in the DLLD architecture. Low-quality data is eliminated using the BDD. In order to identify the co-occurrence dependency and inconsistency in the power flow data caused by prospective attacks, the following CNN is used as a multiple-label classifier. FL uses a central server to compile all client updates. This can cause snags and a decline in performance, particularly when there are a lot of clients. Furthermore, communication delays may arise since each iteration requires all clients to communicate with the central server, particularly for distant clients or crowded networks. In a nutshell, the main contributions of this paper are summarized as follows.

\begin{enumerate}
    \item We propose a privacy-preserved FDI attack detection method over edge-based smart metering networks, by developing an efficient FL framework across distributed edge computing servers. Our method allows for training a global FDI attack detection model at the grid operator's server with good generalization.
    \item We explicitly analyze our method design and algorithm development. We focus on developing a robust FDI attack detection model without sharing data by having distributed ESs, located closer to customers, running an ML-based FDI attack detection model, and sharing the trained model with the grid operator. Hence, our method can achieve FDI attack detection without data sharing for data privacy preservation. 
    \item We validate our proposed framework using the IEEE 14-bus system, showing that the DLLD achieves an average detection accuracy of 88\%. This demonstrates that federated learning is highly accurate, scalable, and robust.

\end{enumerate}

\subsection{Paper Organization}
The remainder of this paper is organized as follows. We present our power system state estimation architecture in Section II. We present our federated learning (FL) framework in the SMs network with edge computing in Section III. In Section IV, we present the simulation and evaluation of our proposed method compared with the existing methods. Section VI concludes the paper. Key acronyms used in this paper are summarized in Table~\ref{tab1}.

\begin{table}[htbp]
\caption{List of key acronyms}
\begin{center}
\begin{tabular}{|c|c|}
% \hline
% \textbf{Table}&\multicolumn{3}{|c|}{\textbf{Table Column Head}} \\
% \cline{2-4} 
% \textbf{Head} & \textbf{\textit{Table column subhead}}& \textbf{\textit{Subhead}}& \textbf{\textit{Subhead}} \\
\hline
Acronym&Definition\\
\hline
FL   & Federated Learning\\
\hline
 ES &   ES\\
 \hline
 BDD & Bad Data Detector\\
 \hline
 CNN &  Convolutional Neural Networks\\
 \hline
 DLLD    & Deep Learning based Locational Detection\\
 \hline
 SM &   Smart Meter\\
 \hline
 FDIA   & False Data Injection Attack\\
 \hline
 ML&  ML\\
 \hline
 SCADA    & Supervisory Control and Data Acquisition\\
 \hline
 EMS &   Energy Management Systems\\
 \hline
 AC   & Alternating Current\\
 \hline
 DC&  Direct Current\\
 \hline
 WLS    & Weighted Least Squares\\
 \hline
 SGD &   Stochastic Gradient Descent\\
 \hline
%\multicolumn{4}{l}{$^{\mathrm{a}}$Sample of a Table footnote.}
\end{tabular}
\label{tab1}
\end{center}
\end{table}

\section{System Model}

\begin{figure}[htbp]
\centerline{\includegraphics[width=1.0\linewidth]{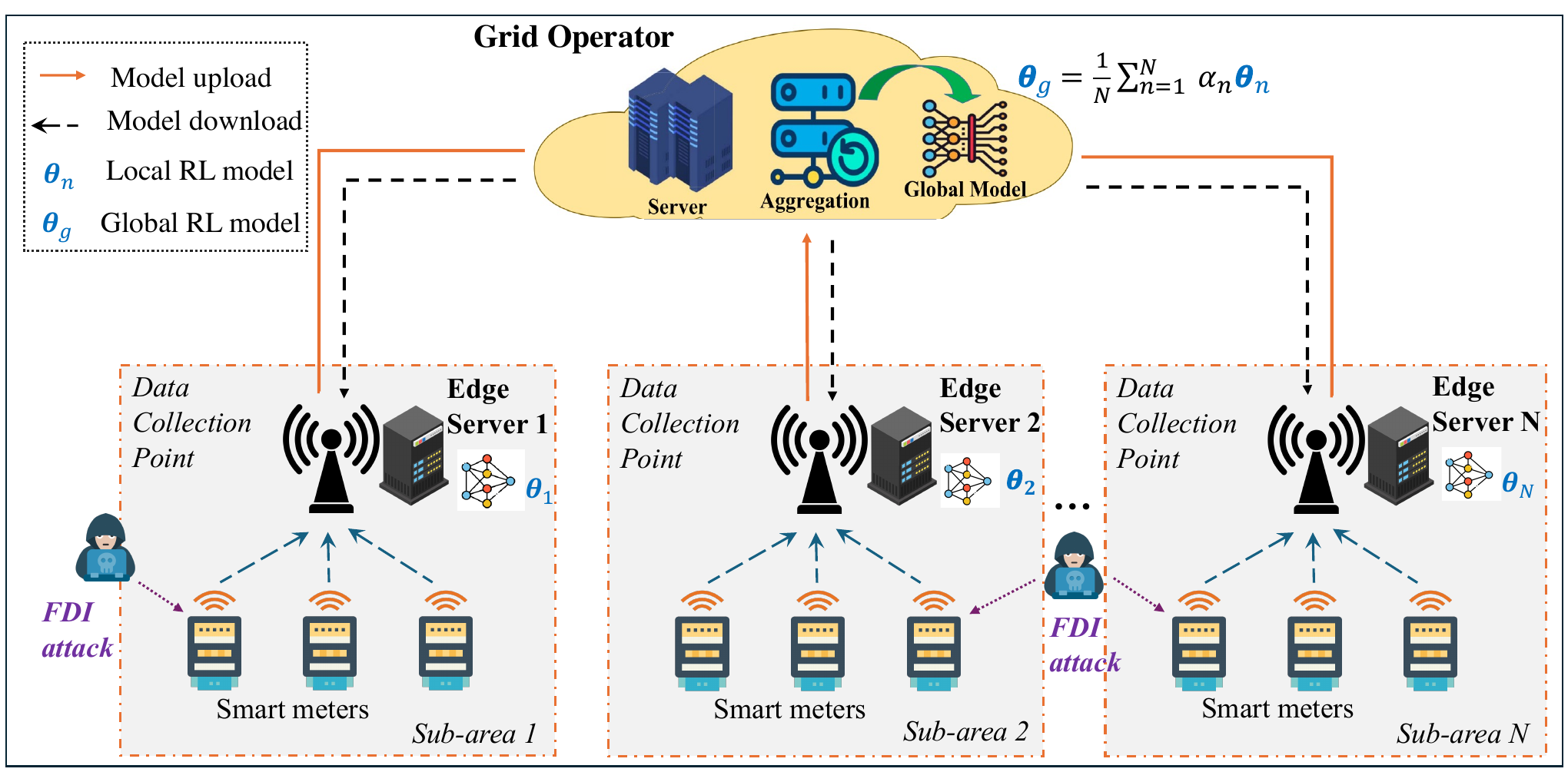}}
\caption{Overview of the Proposed FDI Attack Detection Architecture.}
\label{fig:overview}
\end{figure}

In this paper, we consider a FDI attack detection model shown in Fig. \ref{fig:overview}, consisting of a grid operator's server and a netnetwork of edge servers (ESs). Each ES manages a group of SMs in its coverage area and implements FDI attack detection on behalf of its SMs. The set of ESs is represented by $\mathcal{M}=\{1, 2, \dots,M\}$. We consider that the FL model training takes place over a number of global rounds, denoted by $\mathcal{K}=\{1, 2, \dots,K\}$. In this regard, in each global round $k$, an ES $m$ runs an ML-based FDI attack detection for its group of SMs and shares its trained model update to the grid operator for model aggregation. The ultimate goal is to build a global FDI attack detection model with good generalization during the training stage, aiming to effectively detect FDI attacks in the inference stage.

% Our proposed FDI attack detection architecture is shown in Fig. \ref{fig:overview}. Each ES is positioned to offer computational services to multiple SMs simultaneously. Here, we consider SMs denoted as $N$. This architecture also includes a central grid operator that manages a global model, $\theta_g$, by aggregating local model updates, $\theta_i$, from multiple ESs located across various sub-areas. Each ES, responsible for a specific sub-area, collects data from SMs and trains a local reinforcement learning model. These servers periodically upload their local model updates to the grid operator, which integrates these updates to enhance the global model. The updated global model is then disseminated back to the ESs for implementation, facilitating continuous improvement in FDI attack detection while maintaining data privacy and minimizing communication overhead.

\subsection{Power System State Estimation}
In FDI attacks, the attacker tries to modify the state estimation recorded by SMs. The process of state estimation involves determining a power system's operating condition from the available meter measurements. The state estimation is a crucial analytical technique for accurate power system monitoring. It offers feedback on the network's operating state using measurement data and network information \cite{pires2023dc}.  The state estimation of Alternating Current (AC) is widely used in smart grids. This method makes use of a nonlinear relationship between the system states and the measurements. The AC state estimation model is expressed as follows \cite{pires2023dc}:

% in a supervisory control and data acquisition (SCADA) system. We focus on state estimation in DC microgrids, which are more commonly used than AC ones because of their many benefits \cite{pires2023dc}. These benefits include increased reliability, easier control, and more effective interfacing with energy storage units and renewable energy sources.

%  state estimation is a crucial analytical technique for accurate power system monitoring. It offers feedback on the network's operating state using measurement data and network information. 

\begin{equation}
y = f(v) + d,
\end{equation}
where $y = \{y_1, y_2, \ldots, y_I\}$ represents the measurement vector, which includes the power injection at each bus and the power flow across each branch; $v = \{v_1, v_2, \ldots, v_J\}$ denotes the state vector, indicating the voltage amplitude and phase angle at each bus; $f(v)$ is the measurement function mapping the state vector; $d = \{d_1, d_2, \ldots, d_I\}$ symbolizes the measurement noise; $I$ and $J$ are the number of measurement variables and state variables, respectively.
Various methods for determining the system state vectors, $v$, have been suggested in \cite{khan2021smart}. The most widely employed method in power systems is the Weighted Least Squares (WLS). This approach seeks to find the estimate that minimizes the objective function, which is the sum of the squares of the discrepancies between the actual measurement vector $y$ and the estimated state vector $v$. 
The objective of our analysis is to estimate the state vector \( v \) by minimizing the residual between the observed data \( y \) and the model predictions \( f(v) \). This is accomplished by solving the following optimization problem:

\begin{equation}
\hat{v} = \arg\min_v \left[y - f(v)\right]^T W \left[y - f(v)\right],
\end{equation}

where \( y \) is the vector of measurements, \( f(v) \) is the nonlinear function representing the system model, \( W \) is a diagonal matrix of weights reflecting the relative importance or reliability of each measurement, and \( \hat{v} \) represents the estimated state vector that minimizes the weighted sum of squared residuals. This formulation is commonly used in systems where the accuracy of each measurement can vary, allowing the estimation process to give more weight to more reliable measurements.

Throughout the process of collecting remote terminal measurement data and transmitting it to the power control center's database, several sources of random disturbances can induce errors, such as sensor offsets, communication interferences, and human errors. These errors may cause some measurements to deviate significantly from their true values, referred to as `bad data’. Bad data with substantial errors can severely affect the control center operator's judgment by causing noticeable deviations between the calculated state estimates and the actual conditions. In this regard, the principle of residual testing is applied to detect, process, and eliminate bad data. The residual $r$ is defined as:
\begin{equation}
\|r\|^2 = \|y - f(\hat{v})\|^2.
\end{equation}
Assuming that the state vectors are mutually independent and the random measurement error vectors follow a Gaussian distribution with zero mean, it is established that $\|r\|^2$ conforms to a chi-square distribution. To detect bad data, the Euclidean norm of the residual is compared against a threshold value $\tau$, as specified:
\begin{equation}
\|r\|^2 > \tau.
\end{equation}

\subsection{False Data Intrusion Attack (FDIA) Model}
The technique for creating covert (unobservable) FDI attacks on SM networks is described in this section. The creation of FDIA, based on the chosen system state estimation, is effective and does not result in appreciable computing costs, as stated in \cite{zhuang2019false}. Because the used state estimator has a roughly linear connection, the state estimator can be expressed linearly as follows:

\begin{equation}
\tilde{y} = H\tilde{v} + e,
\end{equation}
where $\tilde{y}$ and $\tilde{v}$ denote the measurement and the closed-form estimated vector, respectively. With the introduction of the attack vector $a$, the residual of the compromised measurements $r_a$ can be expressed as:
\begin{equation}
r_a = \tilde{y}_a - H\tilde{v} = \tilde{y} + a - H\left(\tilde{b}v + \left(H^TWH\right)^{-1}H^TWa\right),
\end{equation}
where $\tilde{y}_a$ represents the compromised measurements, given by $\tilde{y}_a = \tilde{y} + a$. If the attack vector is designed as:
\begin{equation}
a = Hc,
\end{equation}
where $c$ refers to the error injected into the system state by an attacker, then following the attack, the compromised measurement residual $r_a$ will match the measurement residual $r$ before the attack:
\begin{equation}
\begin{aligned}
    r_a = &\tilde{y} - H\tilde{b}v + Hc - H\left(\left(H^TWH\right)^{-1}H^TWHc\right)\\&= \tilde{y} - H\tilde{b}v = r.
\end{aligned}
\end{equation}
If the residual $r$ successfully evades detection by the residual test, then the compromised residual $r_a$, containing malicious data, can similarly evade the residual test. Therefore, attackers can effectively execute unobservable attacks using this model. Hence, developing an effective solution to detect such unobservable FDIA is crucial.

\subsection{FL-based FDI Attack Detection Model}
We consider that each ES $m$ runs an ML model to train a local FDI attack detection model. Specifically, each ML model (e.g., a neural network) associated with data point \( i \) has a loss function \( f(w, i) \in \mathbb{R} \), where \( w_m^k \in \mathbb{R}^d \) represents the ML model (e.g., weights of neurons) of ES $m$ at round $k$. 
% Here, we take into account a situation where all ML training takes place at SMs, with each SM either keeping all of its data locally or fully offloading it to an ES. An SM transfers the fully trained ML model to the ES when it keeps its data. At every global aggregation round \( k \), we define the online loss function of each SM \(n \) as follows:
% \begin{equation}
%     F_n^{(k)}(w) = \frac{1}{|D_n^{(k)}|} \sum_{i \in D_n^{(k)}} f(w, i),
% \end{equation}
% where, \( |D_n^{(k)}| \) denotes the number of data points at SM \( n \) during round \( k \).
% and subsequently define the \textit{online global loss} as
% \begin{equation}
%     F^{(k)}(w) = \frac{1}{D^{(k)}} \sum_{n \in N} D_n^{(k)} F_n^{(k)}(w), \quad D^{(k)} = \sum_{n \in N} D_n^{(k)}.
% \end{equation}
% where, \( F^{(k)}(w) \) is the aggregate online loss across all SMs participating in the training round \( k \). Here, \( D^{(k)} \) represents the total number of data points across all SMs during round \( k \), serving as the normalizing factor for the loss contributions from each SM. The term \( D_n^{(k)} \) refers to the number of data points held by SM \( n \) in round \( k \), and \( F_n^{(k)}(w) \) is the online loss function computed locally at SM \( n \). 
Denote \( D_m^{(k)} \) as the dataset size at ES \( m \) at global round \( k \), the loss function at ES \( m \), \( F_m^{(k)}(w) \), is calculated as follows:
\begin{equation}
    F_m^{(k)}(w) = \frac{1}{D_m^{(k)}} \sum_{i \in D_m^{(k)}} f(w, i),
\end{equation}
where, \( f(w, i) \) represents the loss associated with the \( i \)-th data point, and \( w \) denotes the parameter vector of the ML model. This formula calculates the average loss over all data points within \( D_m^{(k)} \), reflecting the performance of the model at ES \( m \) based on the data it received from smart meters during round \( k \).

We now formalize the ML model training procedure at ESs and the grid operator's server. Each model training round \( k \) starts with the broadcast of a global model, \( w^{(k)} \), from one of the ESs. During round \( k \), each ES \( m \) performs \( e_m^{(k)} \) iterations of SGD over its local dataset, which may vary from one ES to another. The evolution of its local model parameters is given by:
\begin{equation}
    w_m^{(k, e)} = w_m^{(k, e-1)} - \frac{\eta_k}{B_m^{(k)}} \sum_{d \in \mathcal{B}_{m,k,e}} \nabla f(w_m^{(k, e-1)}, d),
\end{equation}
where, \( w_m^{(k, e)} \) denotes the parameters of the model at ES \( m \), at iteration \( e \) of round \( k \). \( \eta_k \) is the learning rate in round \( k \). \( B_m^{(k)} \) is the size of the mini-batch used during the SGD iterations at ES \( m \) in round \( k \). \( \mathcal{B}_{m,k,e} \) is the mini-batch of data points used in iteration \( e \) at ES \( m \) during round \( k \). \( \nabla f(w_m^{(k, e-1)}, d) \) is the gradient of the loss function \( f \) with respect to the model parameters, evaluated at the data point \( d \) and the parameter set \( w_m^{(k, e-1)} \).

After the model training process, each ES \( m \) computes its cumulative gradient, which is used to update the global model parameters. The cumulative gradient at ES \( m \) can be expressed as:
\begin{equation}
    \nabla F_m^{(k)} = \frac{(w^{(k)} - w_m^{(k, e_m^{(k)})})}{\eta_k},
\end{equation}
where, \( w^{(k)} \) denotes the global model parameters after the \( k \)-th update. \( w_m^{(k, e_m^{(k)})} \) represents the local model parameters at ES \( m \) after \( e_m^{(k)} \) local iterations. \( \eta_k \) is the learning rate used in the \( k \)-th iteration.

% Similarly, for each SM \( n \) at the \( k \)-th global iteration, the gradient is computed as:
% \begin{equation}
%     \nabla F_n^{(k)} = \frac{(w^{(k)} - w_n^{(k, e_n^{(k)})})}{\eta_k},
% \end{equation}
% where, \( w_n^{(k, e_n^{(k)})} \) represents the local model parameters at SM \( n \) after \( e_n^{(k)} \) local iterations. These equations illustrate how the local updates at each ES and SM are calculated relative to the current global model parameters.

After local FDI attack detection model training, EMs share their trained models to the grid operator's server to form a new version of the global model. Each participating ES computes its local gradients, these gradients are aggregated to update the global model. The global model   built at the grid operator's server at global round  $k$ is computed as:
\begin{equation}
    w^{(k)} = \sum_{m \in M} \frac{|D_m|}{D}  w_m^{(k)},
\end{equation}
where \( w^{(k)} \) represent the global model at global round $k$, \( \eta_k \) is the learning rate used in the \( k \)-th iteration, and  \( M \) denotes the set of all participating nodes, including both ESs. \( |D_m| \) indicates the number of data points handled by the \( m \)-th node. \( D \) is the total number of data points across all nodes.  

\begin{algorithm}
\caption{Proposed FL algorithm}
\begin{algorithmic}[1]
\State \textbf{Initialization:} Initialize global model parameters \( w^{(0)} \) and distribute them across all ESs.

\For{\( k = 1 \) to \( K \)} \Comment{K is the number of federated learning rounds}
    \State \textbf{Broadcast Global Model:} Send \( w^{(k)} \) from the grid operator to all ESs.
    
    \For{each ES \( m \) connected to local SMs}
        \State \textbf{Local Data Collection:} Gather current round data from SMs under ES \( m \).
        \State \textbf{Local Training:}
        \State Perform \( e^{(k)} \) iterations of SGD to detect FDI patterns.
        \State Update local model parameters \( w_m^{(k+1)} \) using local gradients.
        \State Compute local loss: \( F_m^{(k)}(w) = \frac{1}{|D_m^{(k)}|} \sum_{i \in D_m^{(k)}} f(w, i) \).
        \State Apply differential privacy techniques if necessary to protect data.
    \EndFor

    \State \textbf{Secure Aggregation of Updates:}
    \State Collect and securely aggregate updates \( \Delta w_m^{(k)} \) from each ES.
    \State \( \Delta w^{(k)} = \) Aggregate \( w_m^{(k+1)} - w^{(k)} \) across all ESs.
    
    \State \textbf{Global Model Update:}
    \State Update global model at grid operator: \( w^{(k+1)} = w^{(k)} + \eta_k \Delta w^{(k)} \).
    \State Evaluate global detection performance and adjust parameters as necessary.
    
    \State \textbf{Performance Evaluation and Convergence Check:}
    \State Assess detection accuracy against predefined benchmarks.
    \State Check if the model meets the convergence or performance criteria.
    \If{criteria are met}
        \State Break \Comment{Exit the loop if the model is satisfactory}
    \EndIf
    
\EndFor

\State \textbf{Deployment:} Deploy the finalized global model for real-time FDI detection across the network.
\State \textbf{Output:} The deployed global model \( w^{(K+1)} \), ready for operational use in detecting FDI attacks.
\end{algorithmic}
\end{algorithm}

\textit{Algorithm 1} outlines a federated learning-based approach for detecting FDI attacks. With each ES function as a local client to keep sensitive data localized, this approach improves privacy by enabling data processing and model training directly on SMs. The initial global model parameters are distributed to all ESs by the grid operator at the beginning of the procedure. Overseeing local training on its associated SMs, each ES looks for power utilization irregularities that could be signs of an FDI attack. ESs compute their model gradients after local training and securely send them to the grid operator for aggregation. For ongoing updates and real-time FDI attack detection, the modified global model is then broadcast back to the ESs. Scalable and reliable FDI attack detection in smart meter infrastructure is supported by this method, which minimizes data interchange, lowers communication overhead, and enhances detection capabilities.

\section{Simulations and Evaluation}

The training and testing of the measurement data are shown in this section. We will then go into the model algorithms and execution specifics. Additionally examined are robustness of presence and accuracy detection. For this topic, we will employ multilabel classification with Federated Learning (FL) in order to predict several labels against features of power measurements.

\subsection{Dataset Description}
The efficacy of the detection techniques created for geographically targeted FDI attacks is covered in this section. The experimental framework is the IEEE 14-bus system. There are 19 power measuring locations in this system, which are situated on buses or on connecting lines and are where the measurements are significantly correlated. Line meters are arranged according to index, while injection meters have labels that are arranged in ascending order according to bus indices. Details regarding the dataset and the IEEE 14-bus system are provided in \cite{wang2020locational}.
Artificial load adjustments are applied to all buses to supplement the real-world dataset and create a baseline of unaffected data. Compromise data sets are subsequently generated as well. Since unstructured data anomalies are usually easily detected by traditional bad data detectors, a well-structured FDIA was designed for this study to test detection skills. The standard deviation of noise is adjusted to 0.2 to simulate the measurement noise level. There are 100,000 measurement instances in the training dataset. Ten subsets, each with an equal proportion of compromised and normal data, are used to distribute 10,000 instances for validation. The findings are averaged and covered in the results section. Further details regarding the process of creating the data are provided in \cite{zia2023locational}.

\subsection{Implementation Details}
The system used for all simulations has an RTX3080 GPU, 64.0 GB RAM, and a 12th Gen Intel (R) Core(TM) i9-12900K 3.20 GHz processor. MATLAB's MATPOWER is used for data production, while Python, via PyTorch, is used for the federated learning component. Using Keras \cite{grattarola2021graph}, the multilayer perceptron network (MLP), CNN, and FL are built for increased computational speed.

Regarding the model architecture, the learning rate is set dynamically by the \texttt{ReduceLROnPlateau} callback, and the number of layers includes two hidden layers with 128 and 64 neurons, respectively. Each hidden layer uses \texttt{ReLU} activation, \texttt{L2} regularization with a factor of 0.01, batch normalization, and a dropout rate of 0.4. The output layer uses a \texttt{sigmoid} activation function. The optimizer selected is Adam, and the binary cross-entropy is chosen as the loss function.
As for federated learning, the dataset is split among 5 clients, each training locally for 5 epochs per global round. The total number of global communication rounds is set at 100. During training, the learning rate is adjusted using the \texttt{ReduceLROnPlateau} callback, which reduces the learning rate by a factor of 0.1 if the validation loss does not improve for 10 consecutive epochs. \texttt{StandardScaler} is used to normalize the data before training.

\subsection{Detection Performance}

\subsubsection{Performance Evaluation Matrix}
As performance evaluation criteria, we use precision and recall in our experiments. The mathematical definitions of precision and recall are as follows:
\begin{equation}
\text{Precision} = \frac{\text{True Positive Rate}}{\text{True Positive Rate} + \text{False Positive Rate}},
\end{equation}
and
\begin{equation}
\text{Recall} = \frac{\text{True Positive Rate}}{\text{True Positive Rate} + \text{False Negative Rate}}.
\end{equation}
The probability that a compromised location is correctly labeled as compromised, an uncompromised location is incorrectly labeled as compromised, and an uncompromised location is correctly labeled as uncompromised are the definitions of the terms true positive rate (TPR), false positive rate (FPR), and false negative rate (FNR) in this article.\cite{wang2020locational}, \cite{zhang2020detecting}

We also compute the F1-score in an effort to achieve a balance between recall and precision. The harmonic mean of recall and precision, or the F1-score, can be written as follows:
\begin{equation}
\text{F1-Score} = \frac{2 \times \text{Precision} \times \text{Recall}}{\text{Precision} + \text{Recall}}.
\end{equation}
\subsubsection{The IEEE 14-bus System}
 
\begin{figure}[htbp]
\centerline{\includegraphics[width=1.10\linewidth]{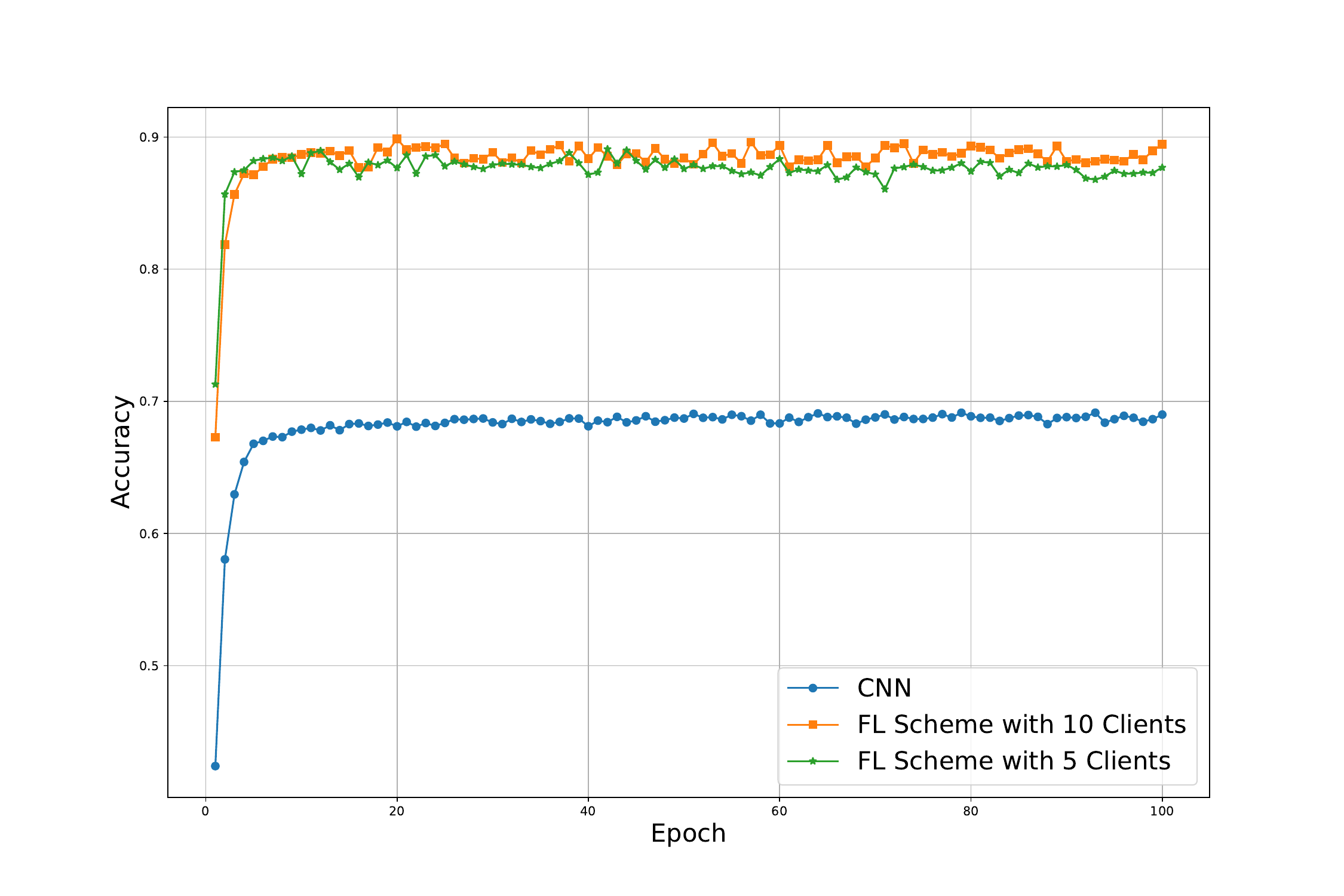}}
\caption{Accuracy comparison between CNN and proposed FL method.}
\label{fig:accuracy}
\end{figure}

The different CNN assessment criteria are displayed in Fig.~\ref{fig:accuracy}, along with two FL schemes that are applied to the IEEE 14-bus system and have client configurations of five and ten, respectively. The accuracy plateau that the CNN model consistently displays at roughly 69\% indicates its capacity limitations or early convergence within the parameters of the dataset and architecture. On the other hand, the FL designs demonstrate far higher accuracy; the 10-client system achieves almost 90\%, while the 5-client scheme achieves approximately 87\%. Interestingly, the 10-client FL scheme not only achieves higher accuracy but also shows lower variance in its performance metrics, indicating improved generalization and stability of the model because of the combined volume and diversity of input from a larger number of customers. The FL approach is effective for large distributed systems, such as smart grids, where data security and integrity are critical.
\begin{figure}[htbp]
\centerline{\includegraphics[width=1.10\linewidth]{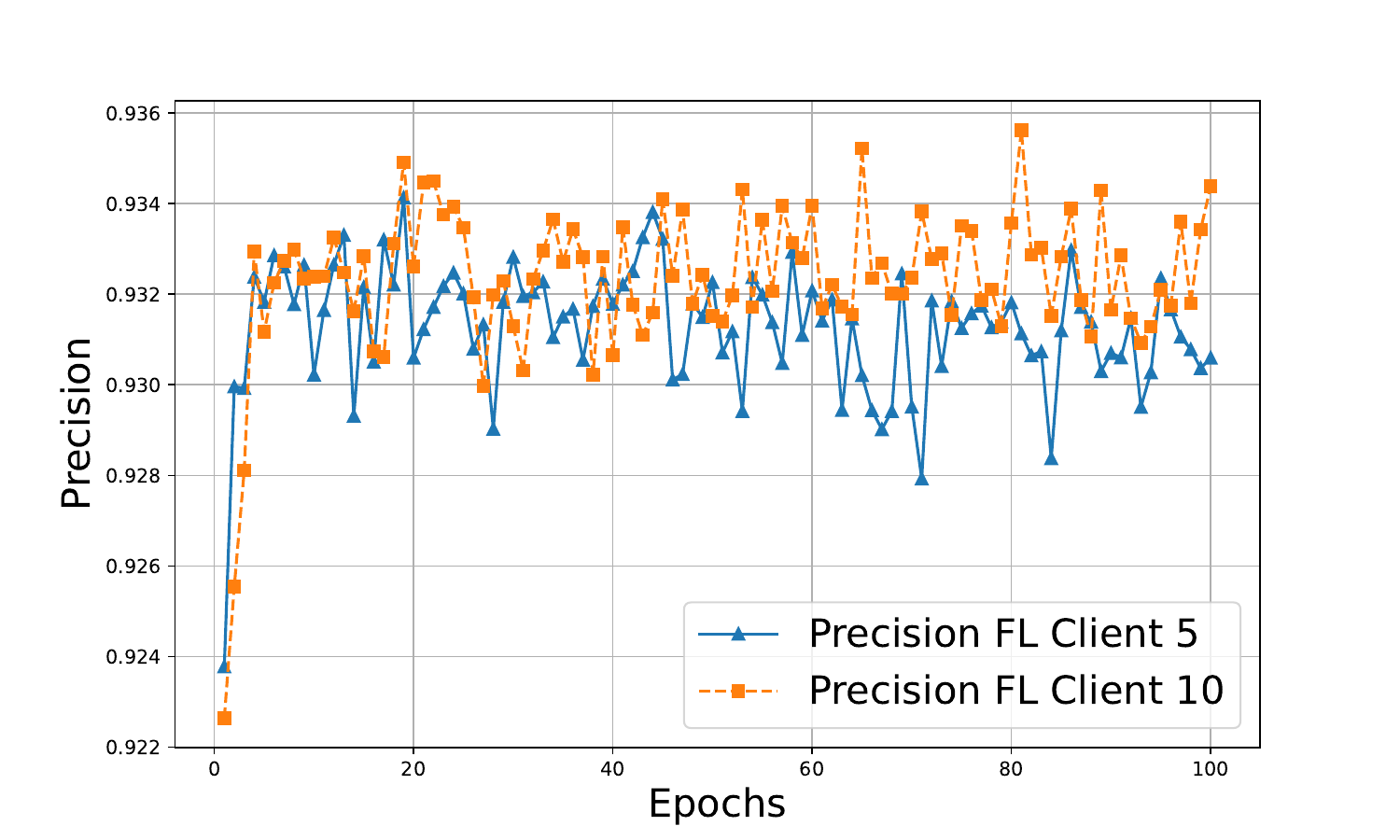}}
\caption{FL performance for precision.}
\label{fig:precision}
\end{figure}
The precision graph for the two Federated Learning (FL) configurations, as depicted in Fig.~\ref{fig:precision}, showcases Client 10 achieving slightly higher precision than Client 5, with precision values generally ranging from 0.930 to 0.936, despite noticeable fluctuations across the epochs.

\begin{figure}[htbp]
\centerline{\includegraphics[width=1.10\linewidth]{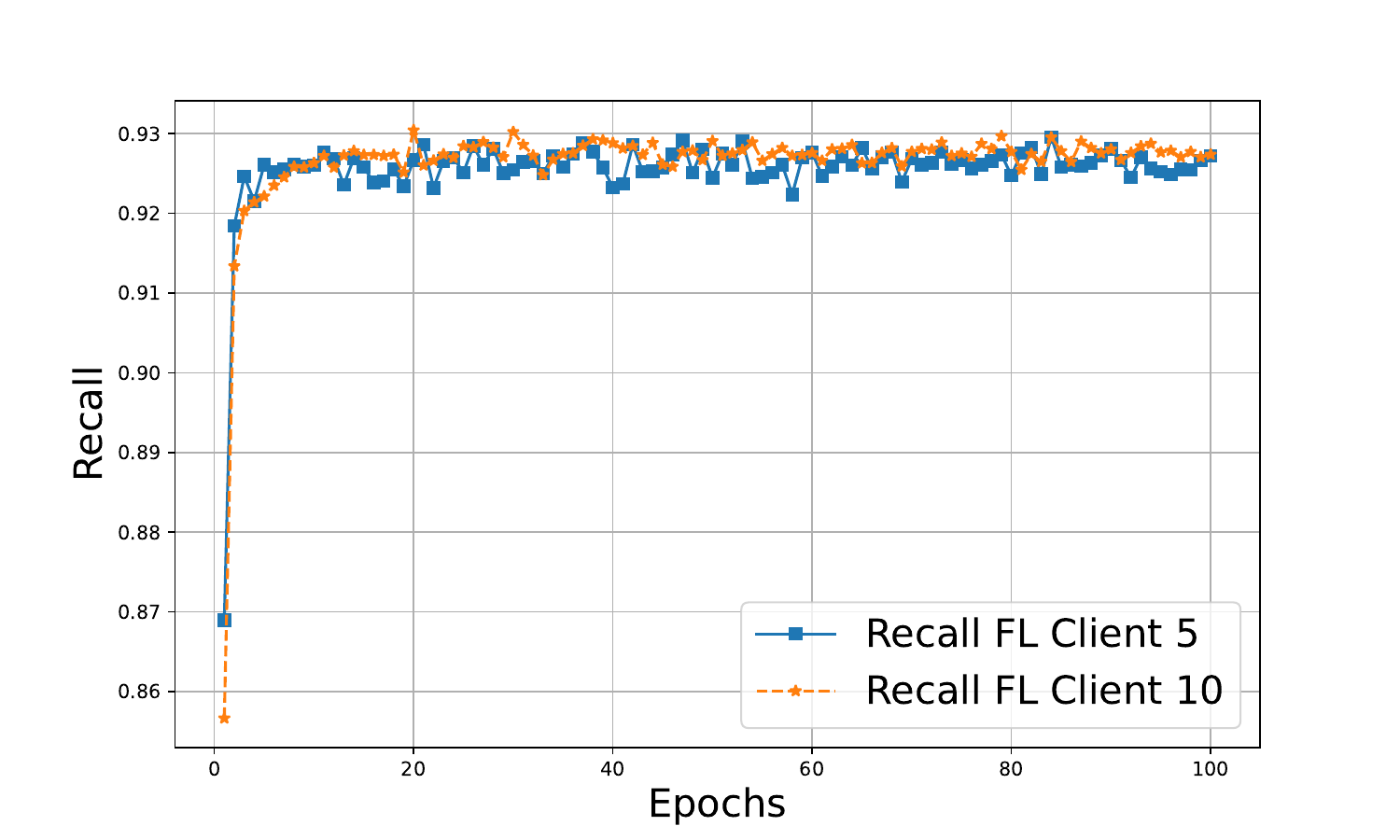}}
\caption{FL performance for recall.}
\label{fig:recall}
\end{figure}

\begin{figure}[htbp]
\centerline{\includegraphics[width=1.10\linewidth]{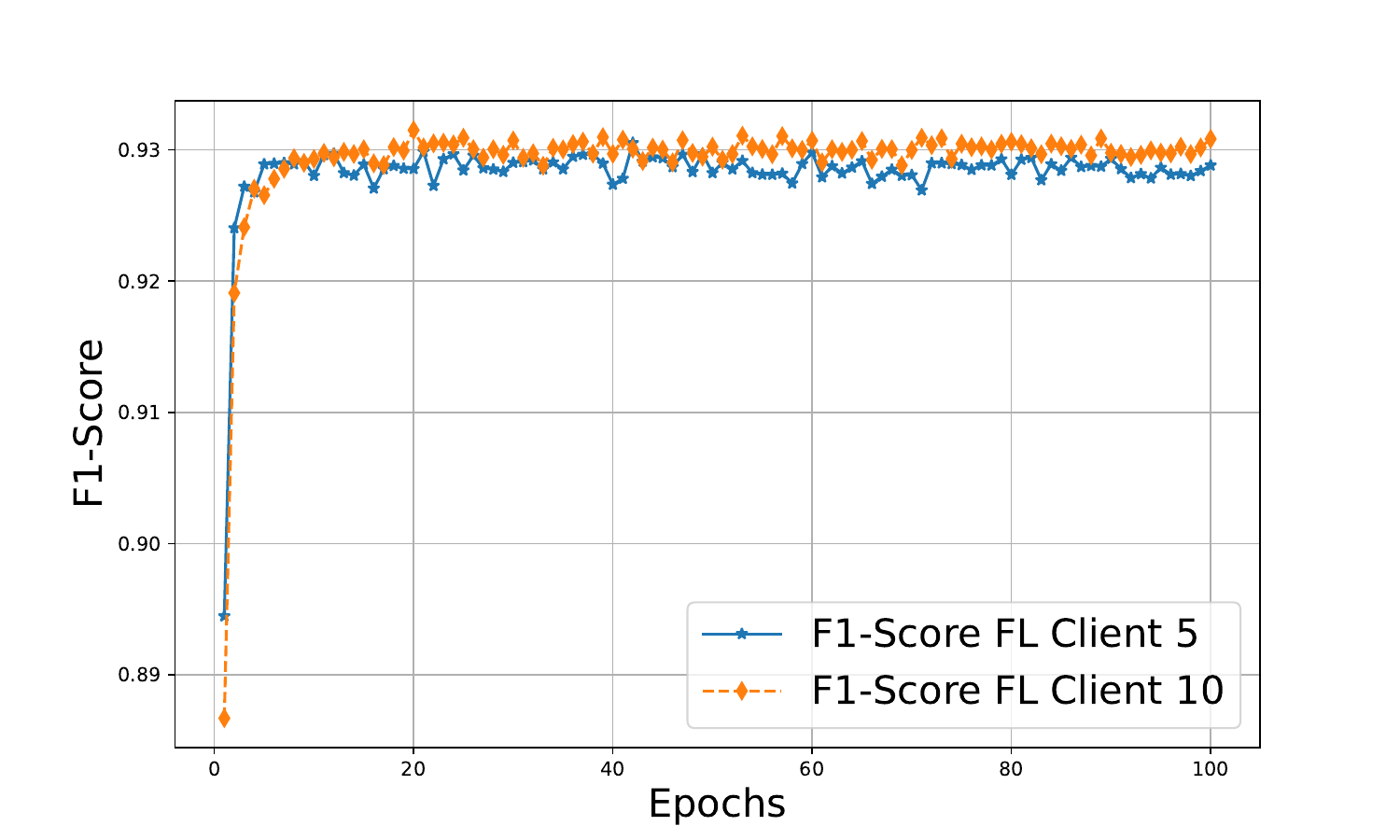}}
\caption{FL performance for F1-Score.}
\label{fig:f1score}
\end{figure}

In contrast, the recall performance illustrated in Fig.~\ref{fig:recall} exhibits an impressive stabilization and consistently maintains high values above 0.90 for both configurations, peaking at approximately 0.93, which demonstrates their robust effectiveness in capturing all relevant instances accurately.
Similarly, the F1-Score, detailed in Fig.~\ref{fig:f1score}, solidifies this observation by maintaining a stable and high performance around 0.92, indicating a well-balanced trade-off between precision and recall.
All of these metrics together demonstrate how much more accurate and dependable the FL models are, as well as how well they can handle false positives and false negatives. This proves that the system performs better overall across a range of operating scenarios.

% \begin{figure*}
%   \centering
%   \begin{tabular}{ c @{\hspace{2pt}} c @{\hspace{2pt}} c}
%     \includegraphics[width=.66\columnwidth]{image/flmlc_precision.pdf} &
%     \includegraphics[width=.66\columnwidth]{image/flmlc_recall.pdf} &
%       \includegraphics[width=.66\columnwidth]{image/flmlc_f1_score.pdf} \\
%     \small (1) FL Performance for Precision.  &
%       \small (2) FL Performance for Recall &
%       \small (3) FL Performance for F1-Score
%   \end{tabular}
%   \medskip
%   \vspace{-5mm}
%   \caption{Comparison of Precision, Recall, and F-1 Score in IEEE 14-Bus System for FL Clients 5 and 10. }
%   \label{fig: ovp}
%   \vspace{-5mm}
% \end{figure*}

\section{Conclusions }
This paper introduces a novel FL framework to detect FDI attacks by developing an efficient federated learning (FL) framework in the SMs network with edge computing. Major concerns about user privacy and data security in smart grids are addressed by this, as it decentralizes the detection process and greatly improves data privacy while reducing the need for extensive data sharing across the network. We demonstrate significant improvements in detection capabilities by precisely identifying attack locations using a multilabel CNN-based FL framework, which is a novel approach. Our methodology is effective based on simulation results using the IEEE 14-bus system, where our FL framework achieved an impressive 88\% average detection accuracy, which is 20\% more accurate than existing methods. This performance is much better than conventional detection techniques, highlighting how federated learning can revolutionize cybersecurity protocols in smart grids. Future work could involve expanding the application of this model to larger grid systems and integrating additional security protocols to fortify defenses against an increasingly sophisticated landscape of cyber threats.
\bibliographystyle{ieeetr}
\bibliography{reference}

\begin{thebibliography}{10}

\bibitem{esmalifalak2014detecting}
M.~Esmalifalak, L.~Liu, N.~Nguyen, R.~Zheng, and Z.~Han, ``Detecting stealthy false data injection using machine learning in smart grid,'' {\em IEEE Systems Journal}, vol.~11, no.~3, pp.~1644--1652, 2014.

\bibitem{lin2024privacy}
W.-T. Lin, G.~Chen, and X.~Zhou, ``Privacy-preserving federated learning for detecting false data injection attacks on power system,'' {\em Electric Power Systems Research}, vol.~229, p.~110150, 2024.

\bibitem{zhao2021federated}
L.~Zhao, J.~Li, Q.~Li, and F.~Li, ``A federated learning framework for detecting false data injection attacks in solar farms,'' {\em IEEE Transactions on Power Electronics}, vol.~37, no.~3, pp.~2496--2501, 2021.

\bibitem{he2017real}
Y.~He, G.~J. Mendis, and J.~Wei, ``Real-time detection of false data injection attacks in smart grid: A deep learning-based intelligent mechanism,'' {\em IEEE Transactions on Smart Grid}, vol.~8, no.~5, pp.~2505--2516, 2017.

\bibitem{ayad2018detection}
A.~Ayad, H.~E. Farag, A.~Youssef, and E.~F. El-Saadany, ``Detection of false data injection attacks in smart grids using recurrent neural networks,'' in {\em 2018 IEEE power \& energy society innovative smart grid technologies conference (ISGT)}, pp.~1--5, 2018.

\bibitem{lin2022incentive}
W.-T. Lin, G.~Chen, and Y.~Huang, ``Incentive edge-based federated learning for false data injection attack detection on power grid state estimation: A novel mechanism design approach,'' {\em Applied energy}, vol.~314, p.~118828, 2022.

\bibitem{tran2023efficient}
H.-Y. Tran, J.~Hu, X.~Yin, and H.~R. Pota, ``An efficient privacy-enhancing cross-silo federated learning and applications for false data injection attack detection in smart grids,'' {\em IEEE Transactions on Information Forensics and Security}, vol.~18, pp.~2538--2552, 2023.

\bibitem{wang2020locational}
S.~Wang, S.~Bi, and Y.-J.~A. Zhang, ``Locational detection of the false data injection attack in a smart grid: A multilabel classification approach,'' {\em IEEE Internet of Things Journal}, vol.~7, no.~9, pp.~8218--8227, 2020.

\bibitem{pires2023dc}
V.~F. Pires, A.~Pires, and A.~Cordeiro, ``Dc microgrids: Benefits, architectures, perspectives and challenges,'' {\em Energies}, vol.~16, no.~3, p.~1217, 2023.

\bibitem{khan2021smart}
M.~A. Khan and B.~Hayes, ``Smart meter based two-layer distribution system state estimation in unbalanced mv/lv networks,'' {\em IEEE Transactions on Industrial Informatics}, vol.~18, no.~1, pp.~688--697, 2021.

\bibitem{zhuang2019false}
P.~Zhuang, R.~Deng, and H.~Liang, ``False data injection attacks against state estimation in multiphase and unbalanced smart distribution systems,'' {\em IEEE Transactions on Smart Grid}, vol.~10, no.~6, pp.~6000--6013, 2019.

\bibitem{zia2023locational}
M.~F. Zia, U.~Inayat, W.~Noor, V.~Pangracious, and M.~Benbouzid, ``Locational detection of false data injection attack in smart grid based on multilabel machine learning classification methods,'' in {\em 2023 IEEE IAS Global Conference on Renewable Energy and Hydrogen Technologies (GlobConHT)}, pp.~1--5, IEEE, 2023.

\bibitem{grattarola2021graph}
D.~Grattarola and C.~Alippi, ``Graph neural networks in tensorflow and keras with spektral [application notes],'' {\em IEEE Computational Intelligence Magazine}, vol.~16, no.~1, pp.~99--106, 2021.

\bibitem{zhang2020detecting}
Y.~Zhang, J.~Wang, and B.~Chen, ``Detecting false data injection attacks in smart grids: A semi-supervised deep learning approach,'' {\em IEEE Transactions on Smart Grid}, vol.~12, no.~1, pp.~623--634, 2020.

\end{thebibliography}
%\printbibliography

\end{document}